\pdfoutput=1

\documentclass[compsoc, letterpaper, 10 pt, conference]{ieeeconf}
\usepackage[utf8]{inputenc}
\usepackage{amssymb}
\usepackage{amsmath}
\usepackage{mathtools}
\usepackage{commath}
\usepackage{soul}
\usepackage[sc,osf]{mathpazo}

\usepackage[utf8]{inputenc}
\usepackage[english]{babel}
\usepackage{csquotes}
 
\usepackage{csquotes}
\usepackage[
    backend=bibtex,
    style=numeric,
    maxcitenames=2,
    maxbibnames=2,
    giveninits=true
]{biblatex}
\usepackage{xcolor}

\usepackage{graphicx}

\usepackage{epstopdf}
\graphicspath{{../}}

\usepackage[caption=false, font=footnotesize]{subfig}

\usepackage{booktabs}
\usepackage{tabularx}
\usepackage{array}

\begin{document}
%
\title{Performance satisfaction in Harpy,\\ a thruster-assisted bipedal robot}

\IEEEoverridecommandlockouts  
\author{Pravin Dangol$^{1}$, Alireza Ramezani$^{1}$ and Nader Jalili$^{2}$
\thanks{$^{1}$Pravin Dangol and Alireza Ramezai are with the Department of Electrical and Computer Engineering, Northeastern University, Boston, MA 02115
        {\tt\small dangol.p@husky.neu.edu, a.ramezani@northeastern.edu}}%
\thanks{$^{2}$Nader Jalili is with the Department of Mechanical Engineering, University of Alabama, Tuscaloosa, AL 35487
        {\tt\small njalili@eng.ua.edu}}%
}


%

\newcommand{\aramezedit}[1]{{\color{magenta} #1}}
\newcommand{\aramez}[1]{\aramezedit{[ARAMEZ: #1]}}


\maketitle

\begin{abstract}
We will report our efforts in designing feedback for the thruster-assisted walking of a bipedal robot. We will assume for well-tuned supervisory controllers and will focus on fine-tuning the desired joint trajectories to satisfy the performance being sought. In doing this, we will devise an intermediary filter based on the emerging idea of reference governors. Since these modifications and impact events lead to deviations from the desired periodic orbits, we will guarantee hybrid invariance in a robust fashion by applying predictive schemes within a short time envelope during the double support phase of a gait cycle. To achieve the hybrid invariance, we will leverage the unique features in our robot, i.e., the thruster.   



\end{abstract}


%
\IEEEpeerreviewmaketitle

\section{Introduction}

Raibert's hopping robots \cite{raibert1984experiments} and Boston Dynamic's BigDog \cite{raibert2008bigdog} are amongst the most successful examples of legged robots, as they can hop or trot robustly even in the presence of significant unplanned disturbances. Other than these successful examples, a large number of humanoid robots have also been introduced. Honda's ASIMO \cite{hirose2006honda} and Samsung's Mahru III \cite{kwon2007biped} are capable of walking, running, dancing and
going up and down stairs, and the Yobotics-IHMC \cite{5354430} biped can recover from pushes.

Despite these accomplishments, all of these systems are prone to falling over. Even humans, known for natural and dynamic gaits, whose performance easily outperform that of today's bipedal robot cannot recover from severe pushes or slippage on icy surfaces. Our goal is to enhance the robustness of these systems through a distributed array of thrusters. 

\begin{figure}[t!] \label{fig:CAD}
\centering
    \includegraphics[width=0.7\linewidth]{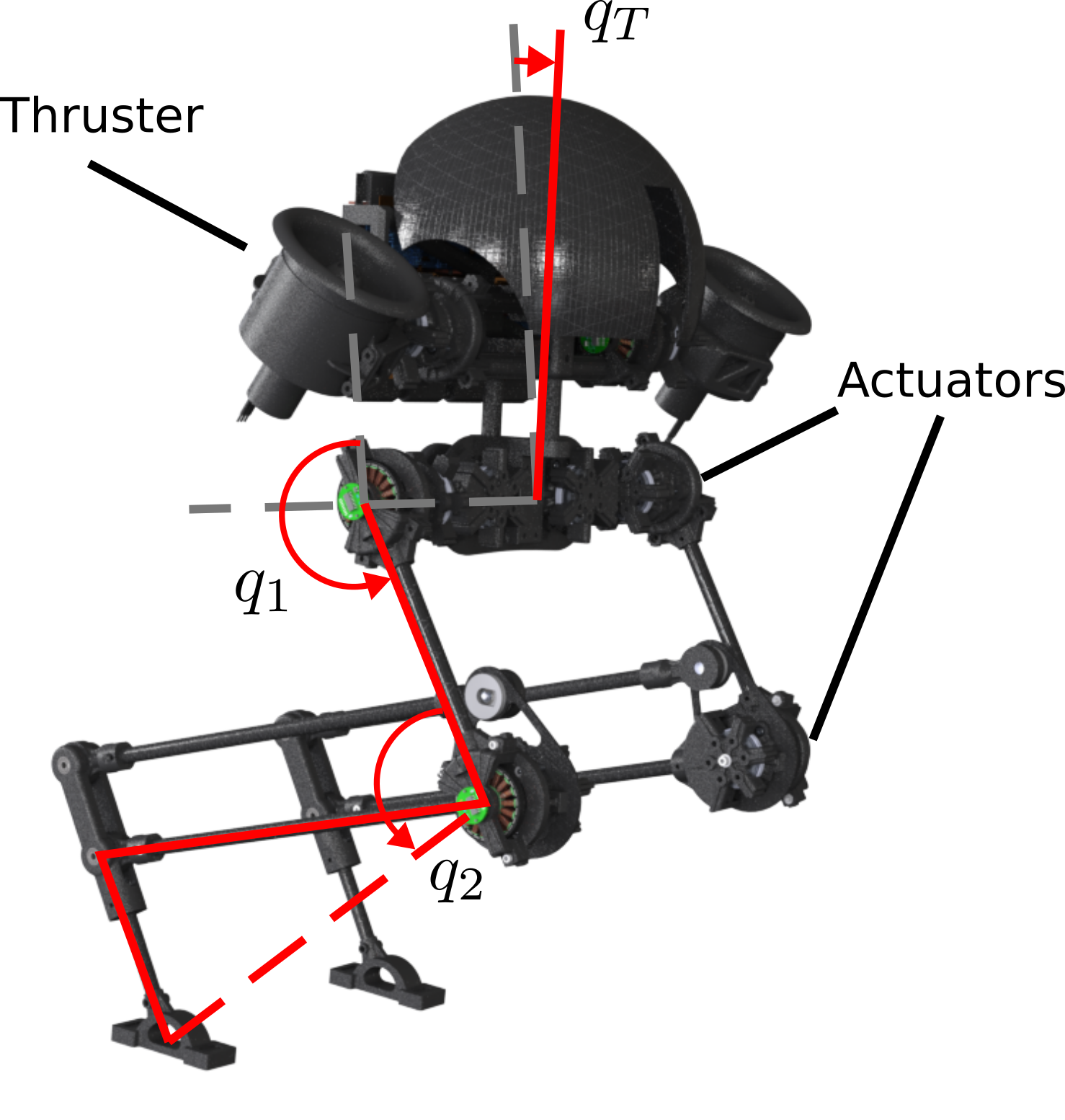}
    \caption{A CAD render of Harpy}
\end{figure}

Here, in this paper, we report our efforts in designing feedback for the thruster-assisted walking of a bipedal robot, called Harpy, currently being developed at Northeastern University. The biped is equipped with a total of six actuators, and two pairs of coaxial thrusters fixed to its torso as shown in figure \ref{fig:CAD}. Each leg is equipped with three actuated joints, the actuators located at the hip allow the legs to move sideways and actuation in the lower portion of the legs is realized through a parallelogram mechanism.

Platforms like Harpy that combine aerial and legged modality in a single platform can provide rich and challenging dynamics and control problems. The thrusters add to the array of control inputs in the system (i.e., adds to redundancy and leads to overactuation) which can be beneficial from a practical standpoint and challenging from a feedback design standpoint. Overactuation demands an efficient allocation of control inputs and, on the other hand, can safeguard robustness by providing more resources. 

The challenge of simultaneously providing asymptotic stability and constraint satisfaction in legged system has been extensively addressed \cite{westervelt2007feedback}. The method of hybrid zero dynamics (HZD) has provided a rigorous model-based approach to assign attributes such as efficiency of locomotion in an off-line fashion. Other attempts entail optimization-based approaches to secure safety and performance of legged locomotion \cite{CLFQP}, \cite{7803333}, \cite{7041347}.

The objective being pursued here is key to overcome a number of limitations in our platform and involves \textit{fast performance constraint and impact invariance satisfaction}. To put it differently, smaller robots have faster dynamics and possess limited actuation power and these prohibitive limitations motivate us to look for motion control solutions that can guarantee asymptotic stability and satisfy performance with minimum computation costs. 

Instead of investing on costly optimization-based scheme in single support (SS) phase, we will assume for well-tuned supervisory controllers as found in \cite{sontag1983lyapunov}, \cite{371031}, \cite{bhat1998continuous} and will instead focus on fine-tuning the desired joint trajectories by implementing an intermediary filter based on the emerging idea of reference governors \cite{411031}, \cite{bemporad1998reference},\cite{gilbert2002nonlinear}, \cite{garone2015explicit} in order to satisfy the performance being sought. Since these modifications and impulsive impact events between gaits lead to deviations from the desired periodic orbits, we will enforce invariance to impact in a robust fashion by applying predictive schemes withing a very short time envelope during the gait cycle, i.e. double support (DS) phase. To achieve hybrid invariance, we will leverage the unique features in our robot, i.e., the thruster.   

This work is organized as follows. In section \ref{section:method} the multi-phase dynamics for a planar walking gait is developed. The SS phase is modeled, and a two point impact map followed by a non-instantaneous DS phase are introduced. In SS phase gaits are first designed based on HZD method, constraints are imposed on an equivalent variable length inverted pendulum (VLIP) model through an explicit reference governor (ERG), the equivalent control action are then mapped back to the full dynamics. During DS phase a nonlinear model predictive control (NMPC) scheme is introduced to ensure performance satisfaction and steer states back to zero dynamics manifold ensuring hybrid invariance. Results are shown in section \ref{section:results}, and we conclude the paper in section \ref{section:conclusion}.

\section{Thruster assisted model with extended double-support phase} \label{section:method}



A full cycle of our model involves consecutive switching between 1) SS phase where only one feet is on the ground, 2) an instantaneous impact map that occur at the end of the SS phase and 3) DS phase where both feet stay in contact with the ground. This model is slightly different form previous works on under-actuated planar bipedal locomotion \cite{898695}, \cite{westervelt2003hybrid}, \cite{chevallereau2004} \cite{choi_grizzle_2005}, \cite{1641816} which assume the double support phase is instantaneous. The extended double support phase will provide a time envelope before the onset of the swing phase for post-impact corrections.


\subsection{SS phase}

\begin{figure}[!ht]
\centering
\null\hfill
    \subfloat[\label{fig:sticks:a}]{
        \includegraphics[width=0.32\linewidth]{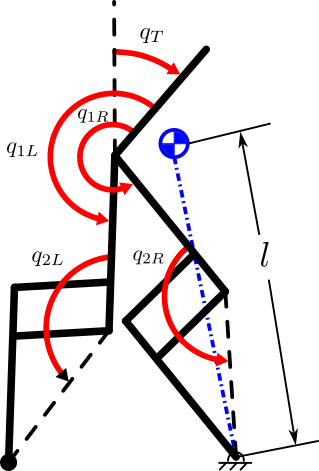}
        }
    \hfill
    \subfloat[\label{fig:sticks:b}]{
        \includegraphics[width=0.3\linewidth]{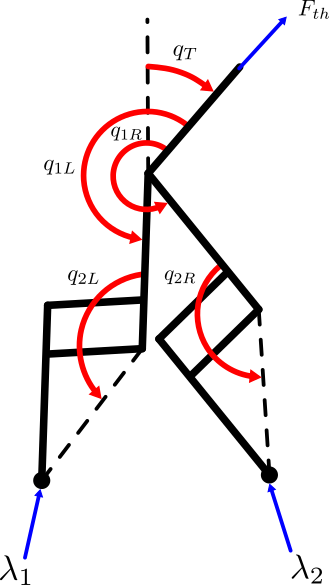}
        }
\hfill\null    
\caption{(a) SS with equivalent VLIP (in blue) and (b) DS models. Dotted line represent the virtual link connecting the feet end to "knee" joint. }
\label{fig:sticks}
\end{figure}

During SS phase the biped has 5 degrees of freedom (DOF), with 4 degrees of actuation (DOA), shown in Fig.~\ref{fig:sticks:a}. Following modeling assumptions widely practiced, it is assumed that the stance leg is fixed to the ground with no slippage, and the point of contact between the leg and ground acts as an ideal pivot. The kinetic $\mathcal{K}(q,\dot{q})$ and potential $\mathcal{V}(q)$ energies are derived to formulate the Lagrangian, $\mathcal{L}(q,\dot{q}) = \mathcal{K}(q,\dot{q}) - \mathcal{V}(q)$, and form the equation of motion \cite{westervelt2007feedback}:
\begin{equation} \label{eqn:swing}
    D_s(q_s)\Ddot{q_s} + H_s(q_s,\dot{q}_s) = B_s(q_s)u 
\end{equation}

\noindent where $D_s$ is the inertial matrix independent of the under-actuated coordinate, $H_s$ matrix contains the Coriolis and gravity terms, and $B_s$ maps the input torques to the generalized coordinates. The configuration variables are as follows: $q_T$ is the absolute torso angle; $q_{1R}$, $q_{1L}$ are the angles of the "femur" relative to torso; and $q_{2R}$, $q_{2L}$ are the angles of virtual "tibia" relative to "femur" as shown in Fig. \ref{fig:CAD} and \ref{fig:sticks:a}. The configuration variable vector is denoted by $q_s = [q_T, q_{1R}, q_{1L}, q_{2R}, q_{2L}]^T \in \mathcal{Q}_s$.


\subsection{Switch between SS and DS}
As pointed out earlier we assume that an impulsive effect similar to what described in \cite{impact} occurs between two continuous modes of SS and DS. We follow the steps from \cite{impact} to model the impact event and solve for the reaction forces $F_{ext}$. The unconstrained version of (\ref{eqn:swing}) are considered by augmenting $q_s$ and including the hip position, $q_e = [q_s, p_H]^T$. The Lagrangian is reformulated and the impulsive force is added on
\begin{equation} \label{eqn:impact}
    D_e(q_e)\Ddot{q_e} + H_e(q_e,\dot{q}_e) = B_e(q_e)u + \delta F_{ext}
\end{equation}

\noindent where the external force $F_{ext}$ acting on each feet end $p=[p_1, p_2]^T$ can be expressed as following
\begin{equation*}
    F_{ext} = J^T \lambda = \begin{bmatrix} 
    \partial p_1 / \partial q_e \\
    \partial p_2 / \partial q_e \\
    \end{bmatrix}^T \lambda
\end{equation*}

\noindent where $\lambda = [\lambda_1, \lambda_2]^T$, shown in Fig.\ref{fig:sticks:b}, is the Lagrange multiplier and assumes that both legs stay on the ground upon impact and Jacobian matrix $J$ is given by $J = \frac{\partial p}{\partial q_e}$. After assuming that the impact is inelastic, angular momentum is conserved and two legs stay in contact with the walking surface, the impact map is resolved  

\begin{equation}  \label{eqn:impact_matrix_ds}
    \begin{bmatrix}
    D_e(q_e^-) & -J(q_e^-)^T \\
    J(q_e^-) & 0_{4 \times 4}
    \end{bmatrix}
    \begin{bmatrix}
    \dot{q}_e^+ \\
    \lambda
    \end{bmatrix}    =
    \begin{bmatrix}
    D_e(q_e^-)\dot{q}_e^- \\
    0_{4 \times 1}
    \end{bmatrix}
\end{equation}


\noindent where the superscript $+$ denotes post-impact and $-$ denotes pre-impact. It is straightforward to show that like \cite{westervelt2007feedback}, the Jacobian matrix $J$ has full row rank and the inertial matrix is always positive definite, the matrix on the left hand side is square and invertible even when both legs are fixed to the walking surface. 






\subsection{Extended DS phase and thrusters}
After impact, both feet stay fixed to the ground. We will assume for a DS phase with constant duration and assume that this duration is significantly smaller than that of the SS phase duration. Legs are swapped, i.e., $q_{R}$ is now $q_{L}$, which is captured by a swapping matrix $R_s^d$ in the following way $[q_{d}, \dot{q}_{d}]^T = R_s^d[q_e^+, \dot{q}_e^+]^T$. The unconstrained dynamics with the ground reaction forces $\lambda$ and the thrusters' action $F_{th}$ are given by 
\begin{equation} \label{eqn:double_support}
    D_{d}(q_{d})\Ddot{q_{d}} + H_{d}(q_{d},\dot{q_{d}}) = B_{d}(q_{d})\eta + J^T \lambda
\end{equation}

\noindent where the control input takes the new form $\eta = [u, F_{th}]^T$. We assume that the relative orientation of the thrust vector with respect to the body stays fixed and is along the torso link. Only changes in the magnitude of the thrust vector are allowed. A damping term (viscous damping) is considered for numerical stability and ease of integration. The kinematics of leg ends are resolved by 
\begin{equation} \label{eqn:no_acc}
    J \Ddot{q}_{d} + \frac{\partial J}{\partial q_{d}} \dot{q}_{d}^2 + d J \dot{q}_{d}= 0 
\end{equation}
\noindent where $d$ is the damping coefficient. The DS dynamical model is captured by the following differential algebraic equation (DAE).  
\begin{equation} \label{eqn:DS_DAE}
    \begin{bmatrix}
    D_{d}(q_{d}) & -J(q_{d})^T\\
    J(q_{d}) &  0_{7 \times 7}
    \end{bmatrix}
    \begin{bmatrix}
    \Ddot{q}_{d} \\
    \lambda
    \end{bmatrix} = 
    \begin{bmatrix}
    B_{d} \eta -  H_{d}(q_{d},\dot{q}_{d}) \\
    -\frac{\partial J(q_{d})}{\partial q_{d}} \dot{q}_{d}^2 - d J \dot{q}_{d}
    \end{bmatrix}
\end{equation}


\subsection{Motion control}
The baseline trajectories are designed according to \cite{westervelt2007feedback}. The restricted dynamics $f_z=f(x_s)+g(x_s)u^*$ on $\mathcal{Z}$, i.e., zero dynamics manifold, is prescribed by the supervisory controller $u^*(x) = - L_g L_f h(x)^{-1} (L^2_f h(x))$ and is invariant of the SS dynamics. This idea is key to HZD-based motion design widely applied to gait design $h_d$ and closed-loop motion control by enforcing holonomic constraint $y = h(x) = q_b - h_d \circ \theta(q)=0$. Where, $q_b = [q_{1R}, q_{1L}, q_{2R}, q_{2L}]^T$ is the vector of actuated coordinates, and $h_d$ is parameteried over the zero dynamics state $\theta(q)$. We applied HZD method to obtain the baseline trajectories for $q_b$ and will take a two-step process including: 1) we will consider the VLIP equivalent model of SS phase and resolve saturated control inputs in a ERG-based framework; 2) We will ensure the gaits are impact invariant by leveraging the thrusters.

\subsection{Explicit reference governor (ERG) and SS phase motion control}
Here, the finite-time enforcement of the holonomic constrained is not our concern and there are a good number of nonlinear control designs for this purpose. With the relative-degree 2, as it is the case here, the feedback linearizing control law $u = L_g L_f h(x)^{-1}(- L_f^2 h(x) + v)$ \cite{khalil2002nonlinear}, where $v = K_P y + K_D \dot{y}$, is one of the simplest options that meets our requirements. Other options are: Control Lyapunov Function based Quadratic Programs \cite{CLFQP}, Sliding Mode Controller \cite{saglam2015meshing}, Passivity based controller \cite{spong2007passivity} to name a few.





\begin{figure}
    \centering
    \includegraphics[width = 0.8\linewidth]{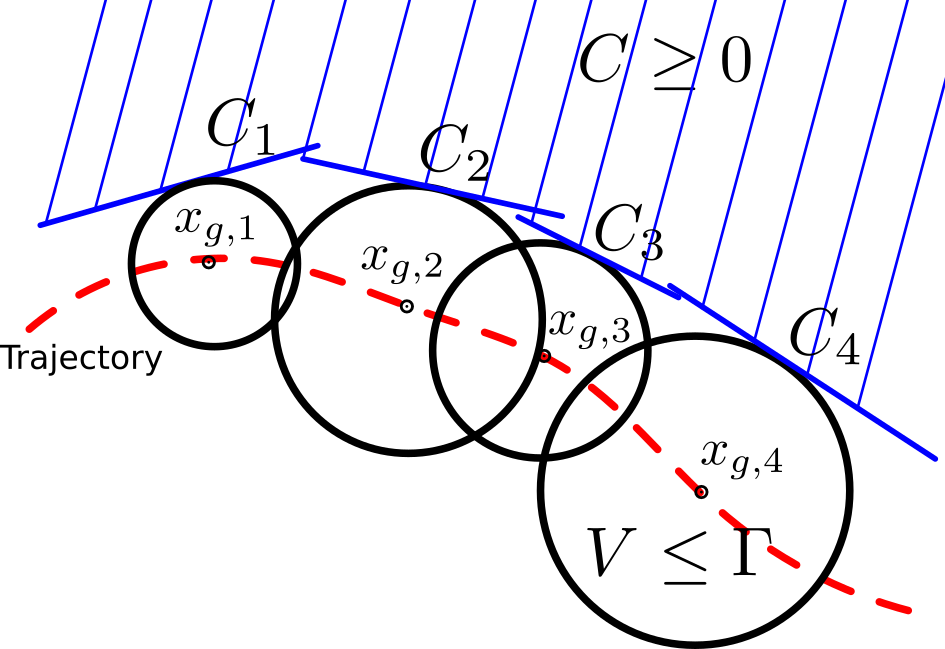}
    \caption{Geometric interpretation of the level set $\{x_{v}|V \leq \Gamma\}$.}
    \label{fig:level set}
\end{figure}

The approach taken here is based on the idea of reference governors \cite{gilbert2002nonlinear} which allows for enforcing the holonomic constraints subject to state and input limits to be separated from the control design. To be more specific an ERG \cite{garone2015explicit} approach is taken to avoid the need for optimization as in \cite{gilbert2002nonlinear}. This separation is nicely defendable after appreciating that the actuation dynamics are very fast (two-time-scale problem). This is not something unusual to assume for high-power actuators typically found in legged robots.  



An equivalent VLIP model for SS phase is considered, which is under-actuated at its base and the variable length $l$ is actuated, shown in figure \ref{fig:sticks:a}. The full control action $u$ in (\ref{eqn:swing}) can be related to the equivalent control $u_v$ \cite{6225272}. The center of mass (COM) trajectory $r$ from HZD model is extracted. The reference governor acts as a supervisory controller that outputs a manipulated reference signal $w$ to ensure that the state and control constraints in the vector $C(x_{v},x_w)$, where the elements of the vector given by 
\begin{align}\label{eqn:c_ERG}
    c_i(x_{v},w) &:= c_{x, i} x_{v} + c_{w, i} x_w + c_{limit, i} \geq 0, \ i = 1,..,n_c
\end{align}
\noindent are satisfied. In (\ref{eqn:c_ERG}), $x_{v} = [l, \dot{l}]^T$, $x_w = [0, w]^T$ is the steady state solution, $c_{limit, i}$ contains limits applied to the states and input. The dynamics of the manipulated reference $w$ is then defined such that the Lyapunov function $V(x_{v}, x_w)$ is bounded by a smooth positive definite function $\Gamma(w)$. 
\begin{align} \label{eqn:V}
    V(x_{v},x_w) \leq \Gamma (w)
\end{align}
The following Lyapunov function is considered 
\begin{align}
    V(x_{v},x_w) = (x_{v} - x_w)^T P (x_{v} - x_w)
\end{align}
\noindent where $P$ is a positive definite matrix consisting of controller gain $K_P$ and equivalent pendulum mass $m_v$ ($P = diag(K_p, m_v) >0$). Geometrically this represents a ball around the steady state solution ($x_w$), a interpretation of this is depicted in Fig. \ref{fig:level set}. The Lyapunov stability argument ($V(x_{v},x_w)$ is positive definite $\forall x_{v} \neq x_w$ and $\dot{V}(x_{v},x_w, \dot{w}) \leq 0$) implies $\{ x_{v} : V(x_{v},x_w) \leq \Gamma(w) \}$ is a positive invariant set, i.e., once $x_{v}$ belongs to this set it converges to $x_w$. 

In order for the constraint in (\ref{eqn:c_ERG}) to be captured in (\ref{eqn:V}) a change of coordinates $\Tilde{x} = P^{1/2}(x_v - x_w)$ is applied which changes the constraint equation to $c_{x,i} P^{-1/2}x_w + c_{x,i}x_w+ c_{w,i}w + c_{limit,i} \geq 0$. Then $\Gamma_i(w)$ is defined as $\Gamma_i(w) = |\Tilde{x}|^2$, which is the squared distance between the constrained and steady state solution $x_w$ \cite{garone2015explicit}:
\begin{align}
    \Gamma_i(w) = \frac{ \big( c_{x, i} x_{w} + c_{w, i} w + c_{limit, i} \big)^2 }{c_{x, i}^T P^{-1} c_{x, i}}
\end{align}

\noindent The upper bound on $V(x_v,x_w)$ is defined as $\Gamma(w) = min (\Gamma_i(w))$, which results in the shortest distance to the boundary formed by $C(x_{v},x_w)$.

The condition given by
\begin{align}\label{eqn:V_dot}
    \dot{V}(x_{v},w,\dot{w}) \leq \dot{\Gamma} (w, \dot{w})
\end{align}
\noindent ensures that the states do not cross the invariant level set. Please note that $\dot{\Gamma}<0$. A continuous reference dynamics is then formulated \cite{garone2015explicit} such that


\begin{align}\label{eqn:dot_w}
    \dot{w} &:= \kappa\left(\Gamma(w) - V(x_{v})\right) \frac{r - w}{||r - w||} sat_1(r-w)
\end{align}
\noindent where $sat_1(\cdot)$ refers to a function that saturates its input between $\pm 1$ and $\kappa$ is an arbitrary large scalar. In (\ref{eqn:dot_w}), $(\Gamma(w) - V(x_{v})$ provides a scaling factor to modify the reference dynamics while the remaining portion of the equation generates an attractive field that allows $w$ to converge to $r$.


The manipulated reference dynamics $\dot{w}$ which satisfies (\ref{eqn:V}) and (\ref{eqn:V_dot}) estimates the nominal reference $r$ as close as possible while satisfying imposed constraints (\ref{eqn:c_ERG}) on the equivalent system.

The control action computer for the equivalent system $u_v$ can then be mapped back to the actual SS model. From the principle of virtual work, the work done in SS phase and its equivalent VLIP model $u^T \delta q_b + u_v \delta l = 0$. Where $l = \sqrt{p_{cm,x}^2 + p_{cm,y}^2}$, then the mapping is:

\begin{align}
    \begin{split}
        u^T = \Upsilon (\tau_l, q) = - u_v \ l^{-1}\Big( \frac{\partial p_{cm,x}}{q_b} + \frac{\partial p_{cm,y}}{q_b} \Big)
    \end{split}
\end{align}

\noindent The contribution from the under-actuated angle $q_T$ is considered to be zero. The equivalent SS phase model is depicted in Fig. \ref{fig:sticks:a}.



\subsection{Impact invariance}
ERG and the two-point impact event causes large deviations from the zero-dynamics manifold and the extended DS phase and thrusters are leveraged to steer the states to the zero dynamics manifold ($\mathcal{Z}$). When DS is absent, hybrid invariance in \cite{westervelt2003hybrid} takes a simpler form ($\Delta(\mathcal{S} \cap \mathcal{Z}) \subset Z$) . Here -- with abuse of notation -- impact invariance $\Pi(\Delta(\mathcal{S} \cap \mathcal{Z})) \subset \mathcal{Z}$ is sought, where $\Pi: x_{d,0} \mapsto x_{s,0}$ maps the initial state of DS phase to the initial state of the subsequent SS phase. Hybrid invariance in this case leads to each gait starting with the same initial condition despite the impulsive effects of impact and deviation from designed trajectories. Please note that for a robot with passive ankles extended DS phase can lead to fall-over and that this system is augmented with thrusters allows to approach this model and seek for stable gaits. 

%
%
%
%
%
%
%
As opposed to the SS phase, the constraints in the DS phase take a more complex form, the ground reaction forces need to be satisfied, the final states at the end of the DS phase ($x_{d,f}$) must match the initial states at the SS phase ($x_{s,0}$) to ensure hybrid invariance. We apply a NMPC-based design scheme to steer the post-DS states back to the zero-dynamics manifold. This scheme is known for being costly, however, the duration of the DS phase is significantly shorter than SS. 

The state-space representation of DS phase $\dot{x}_d= f(x_{d}) + g(x_{d})\eta$, derived from (\ref{eqn:DS_DAE}), where the input vector is augmented to take to $\eta = [u, f_{th}]^T$ is considered for the DS phase.

%
%
%
Note that a reference for each DS state $r_d[k]$ is generated at every k-th sample over the duration of the double support phase. The reference can be a simple linear trajectory between the post-impact states $x_d^+$ and the initial SS phase state $x_{s,0}$. 

The continuous DS phase model is converted into a discrete-time model and is then linearized at each each sample time. The following optimization problem is resolved to minimize the cost function, which is denoted by $\phi (x_d,\eta)$, is given by 

\begin{align} \label{eqn:cost}
    \begin{split}
    \quad & \underset{\eta[k]}{\text{min}}~  \phi (x_d,\eta)   = \sum_{k=1}^{N}  \sum_{i=1}^{p} w_{x,i} ( x_{d,i}[k] - r_{d,i}[k] ) + \\ & \qquad \qquad \qquad \ \ \sum_{k=1}^{N-1} \sum_{j=1}^{m} w_{\eta,j} \Delta \eta_j[k] \\
    & \text{subj. to:} \\ 
    & x_{d}[1] = R_s^d x_e^+\\
    & x_{d}[k+1] = f(x_d[k]) + g(x_d[k])\eta[k]\\
    & \eta_{min} < \eta[k] < \eta_{max} \\
    & x_{d \ min} < x_{d}[k] < x_{d \ max} \\
    & \Big\lvert \frac{\lambda_T[k]}{\lambda_N[k]} \Big\rvert  < \mu_s \\
    & \lambda_N[k] > 0\\
    \end{split}
\end{align}
\noindent where the initial DS phase state $x_{d}[1]$ comes directly from the post impact state $x_e^+$, after the roles of the legs have been swapped which is denoted by $R_s^d$ matrix. The subsequent constraint $x_d[k+1]$ ensures that the discreet linearized states belong to the DS phase. Limits are imposed on both states and control actions through $\eta_{min / max}$ and $x_{d \ min/max}$ respectively. And finally, the ground contact condition must be satisfied for the DS phase i.e., the ratio of tangential $\lambda_T$  to normal forces $\lambda_N$ is less than the static coefficient of friction $\mu_s$ and normal force is always positive. 

With these constrained satisfied, the NMPC guides the DS states towards the initial condition of SS phase, resulting in impact and DS phase invariance.

\begin{figure} 
    \centering
    \includegraphics[width = 0.7\linewidth]{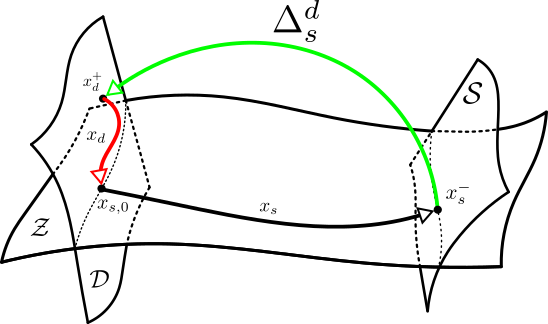}
    \caption{Geometric representation of hybrid zero dynamics and invariance achieved though double support phase $x_{d}$ (red)}
    \label{fig:manifold_DS}
\end{figure}


\section {Numerical \& Experimental Results} \label{section:results}


\begin{figure}
    \centering
    \includegraphics[width=0.7\linewidth]{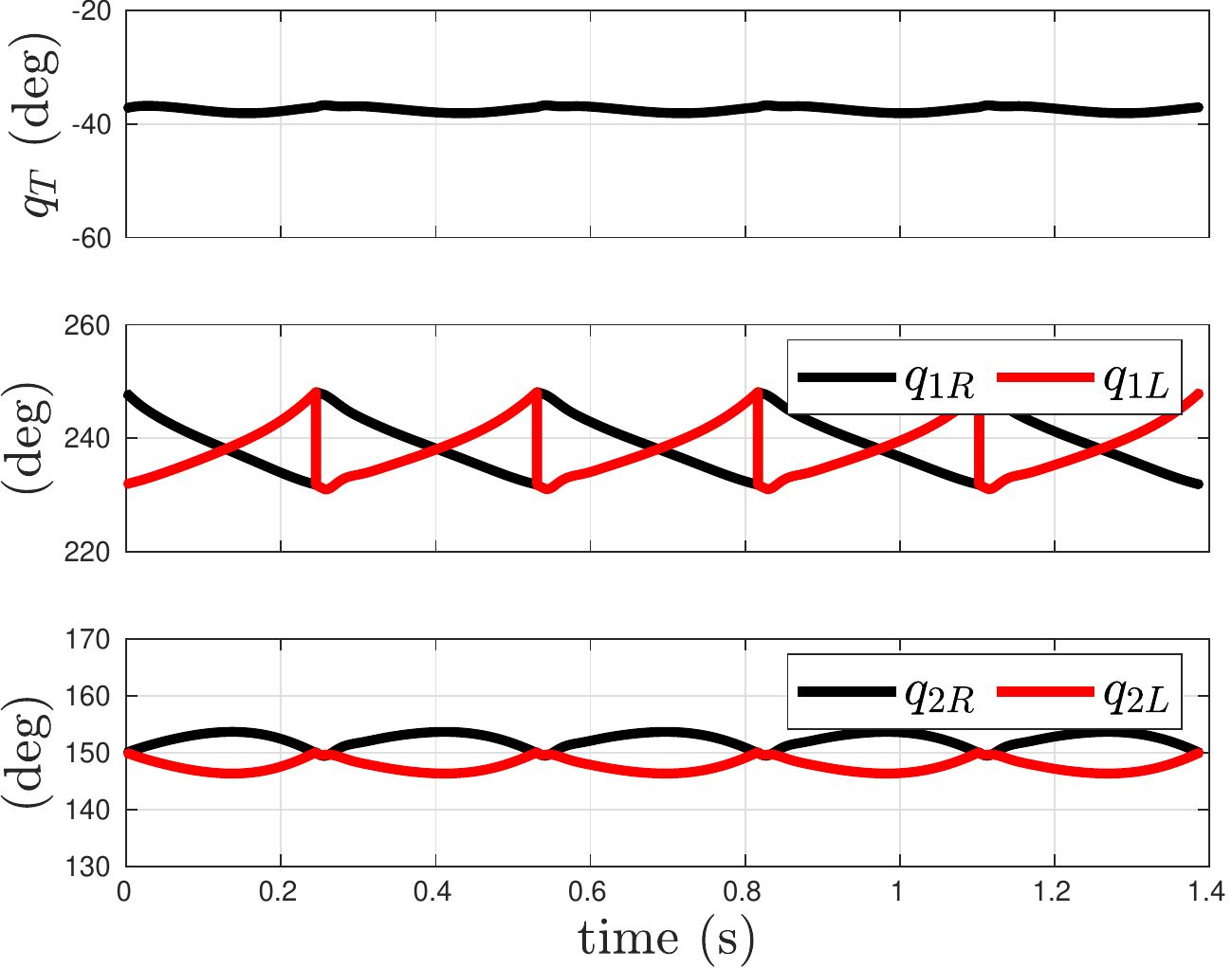}
    \caption{Joint angle trajectories}
    \label{fig:angles}
\end{figure}

\begin{figure}
    \centering
    \includegraphics[width=0.7\linewidth]{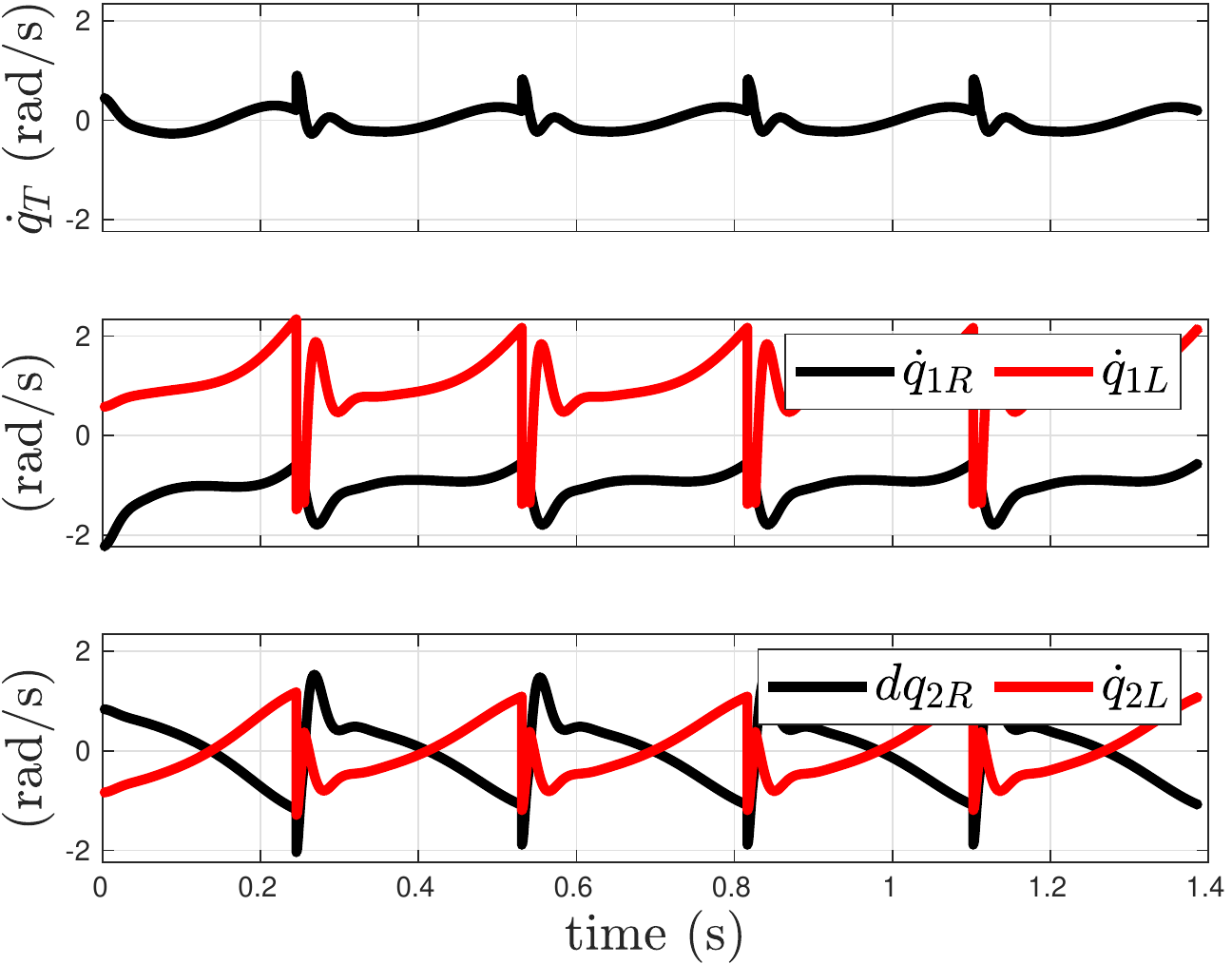}
    \caption{Joint velocity trajectories}
    \label{fig:velocity}
\end{figure}

\begin{figure}
    \centering
    \includegraphics[width=0.8\linewidth]{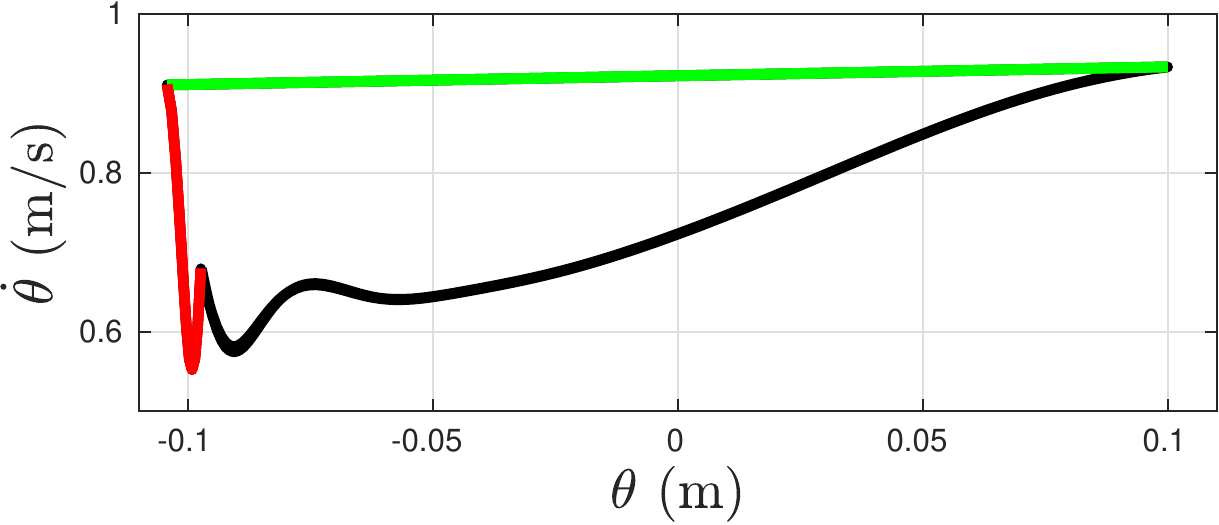}
    \caption{Phase portrait of the zero dynamics states $[\theta,\dot{\theta}]^T$, the black, red and green lines indicate SS, DS and impact, respectively.}
    \label{fig:PP_theta}
\end{figure}

\begin{figure}
\centering
\null\hfill
    \subfloat[Coefficient of friction - leg 1\label{fig:forces:mu1}]{
        \includegraphics[width=0.45\linewidth]{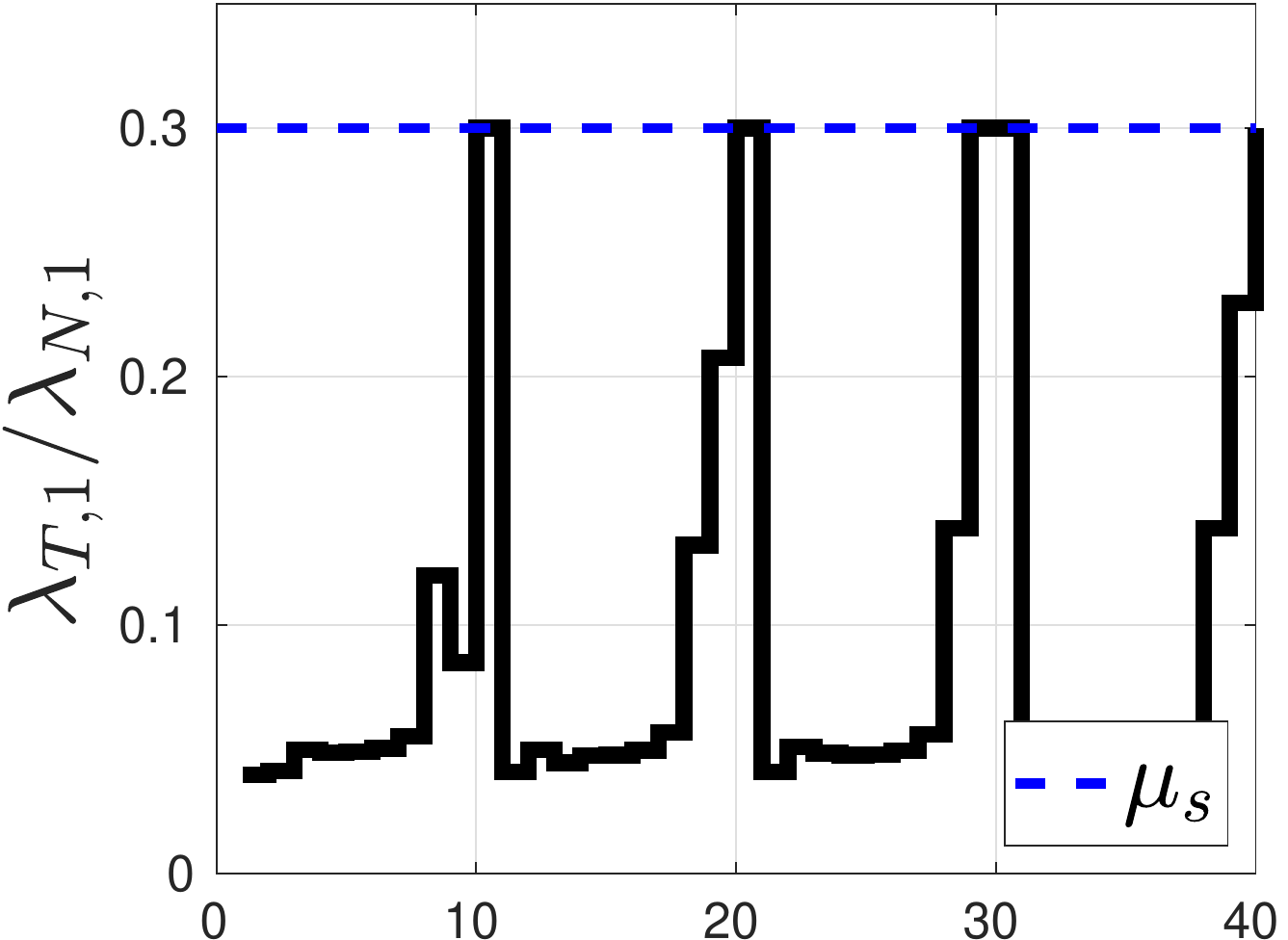} 
        }
    \hfill
    \subfloat[Coefficient of friction - leg 2\label{fig:forces:mu2}]{
        \includegraphics[width=0.45\linewidth]{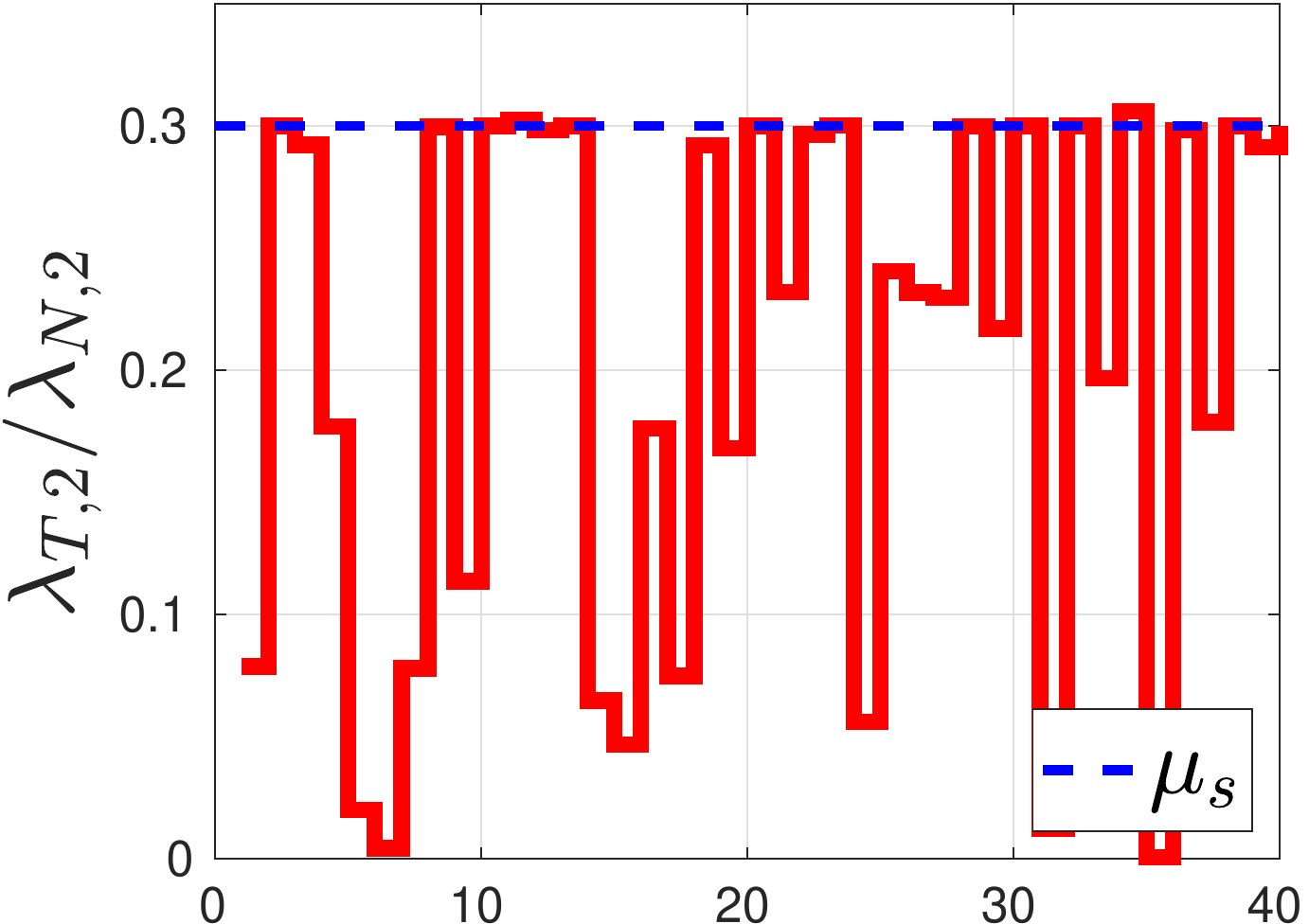}
        }
\hfill\null   
\hfill
\null\hfill
\subfloat[Normal force - leg 1\label{fig:forces:normal1}]{
        \includegraphics[width=0.45\linewidth]{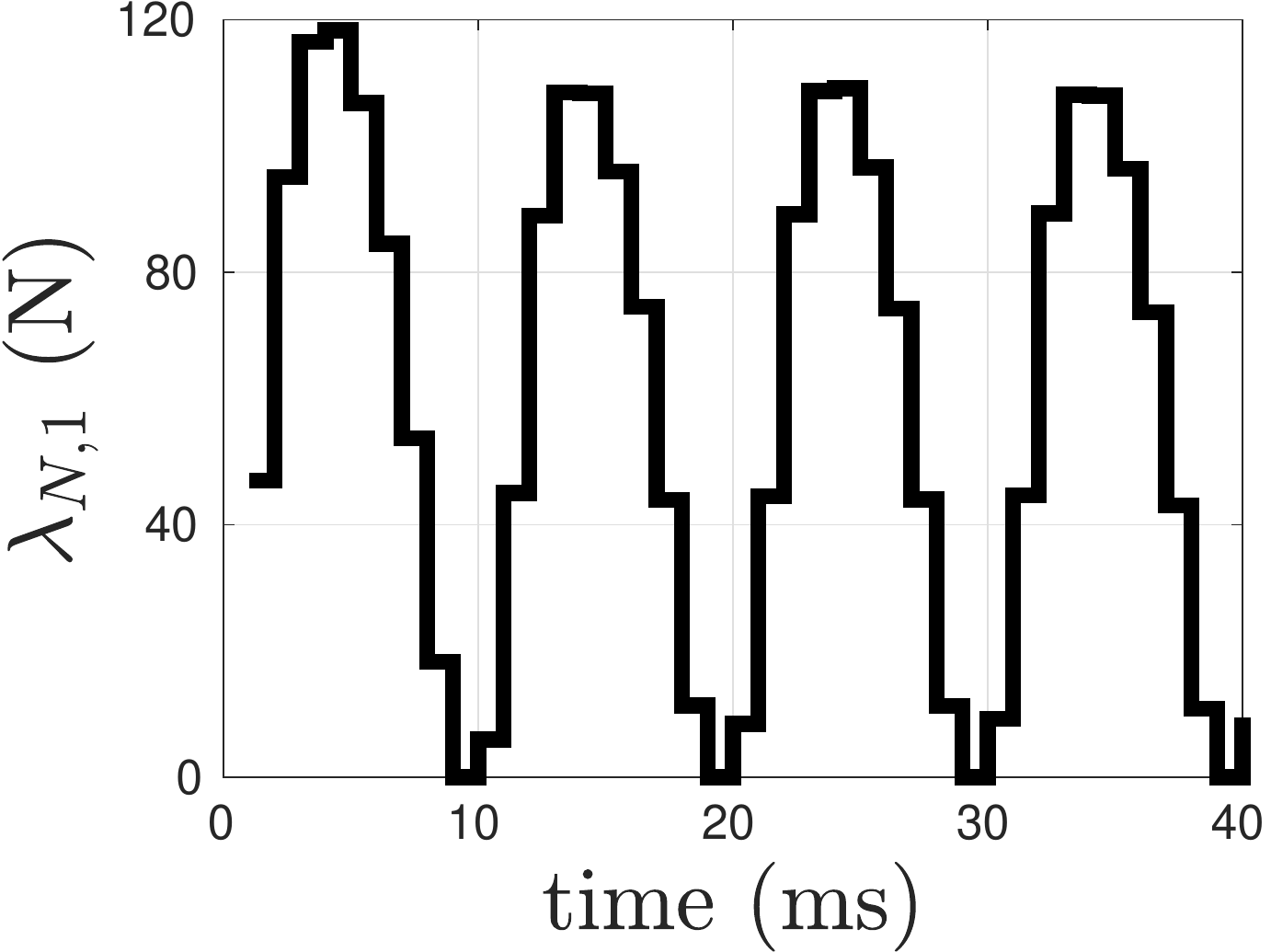}
        }
    \hfill
    \subfloat[Normal force - leg 2\label{fig:forces:normal2}]{
        \includegraphics[width=0.45\linewidth]{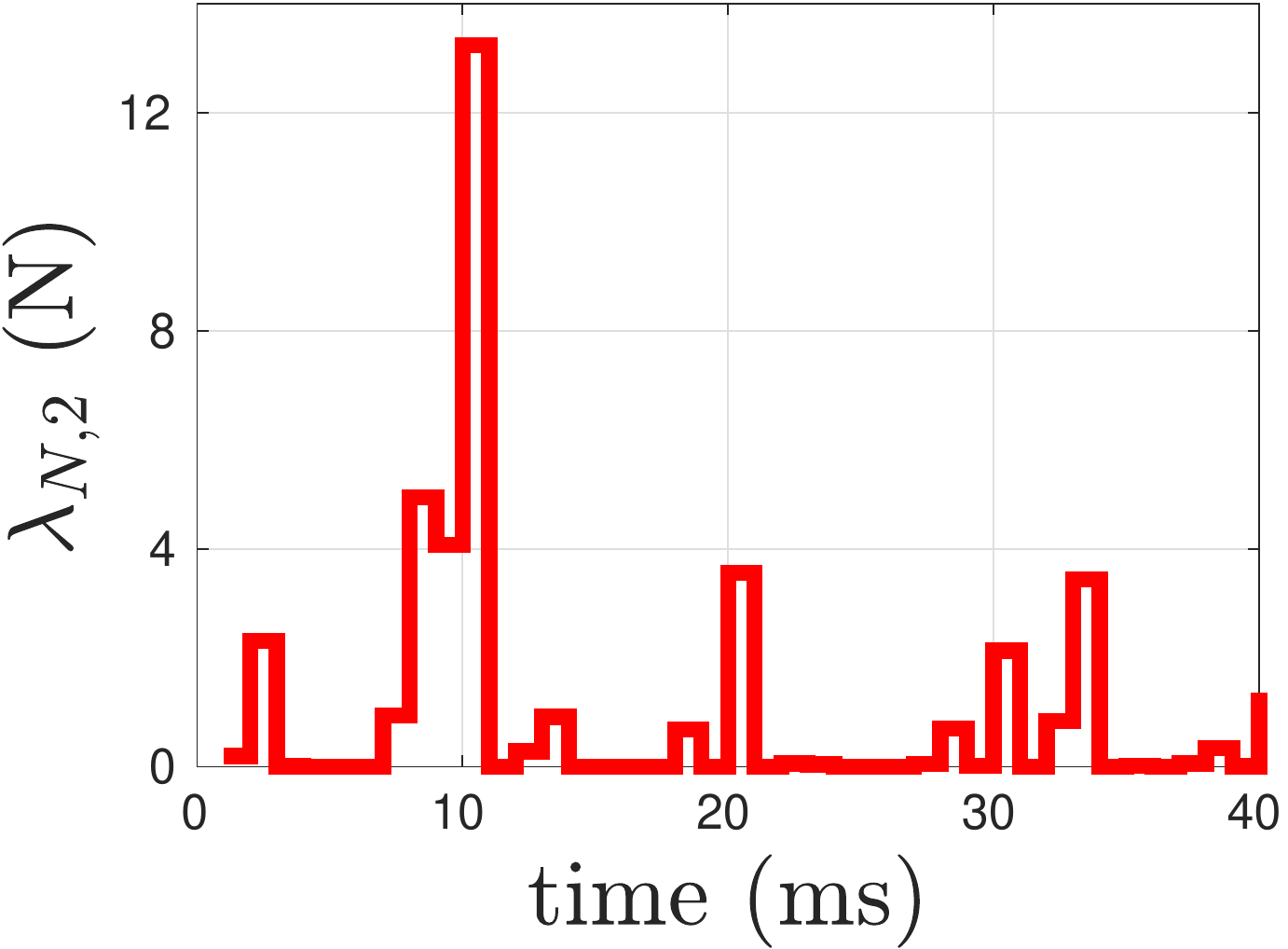}
        }
\hfill\null 
\caption{Friction constraints during DS phase of leg 1 and 2 in black and red respectively. Intermediate SS phases were omitted.}
\label{fig:forces}
\end{figure}

\begin{figure}
    \centering
    \includegraphics[width=0.7\linewidth]{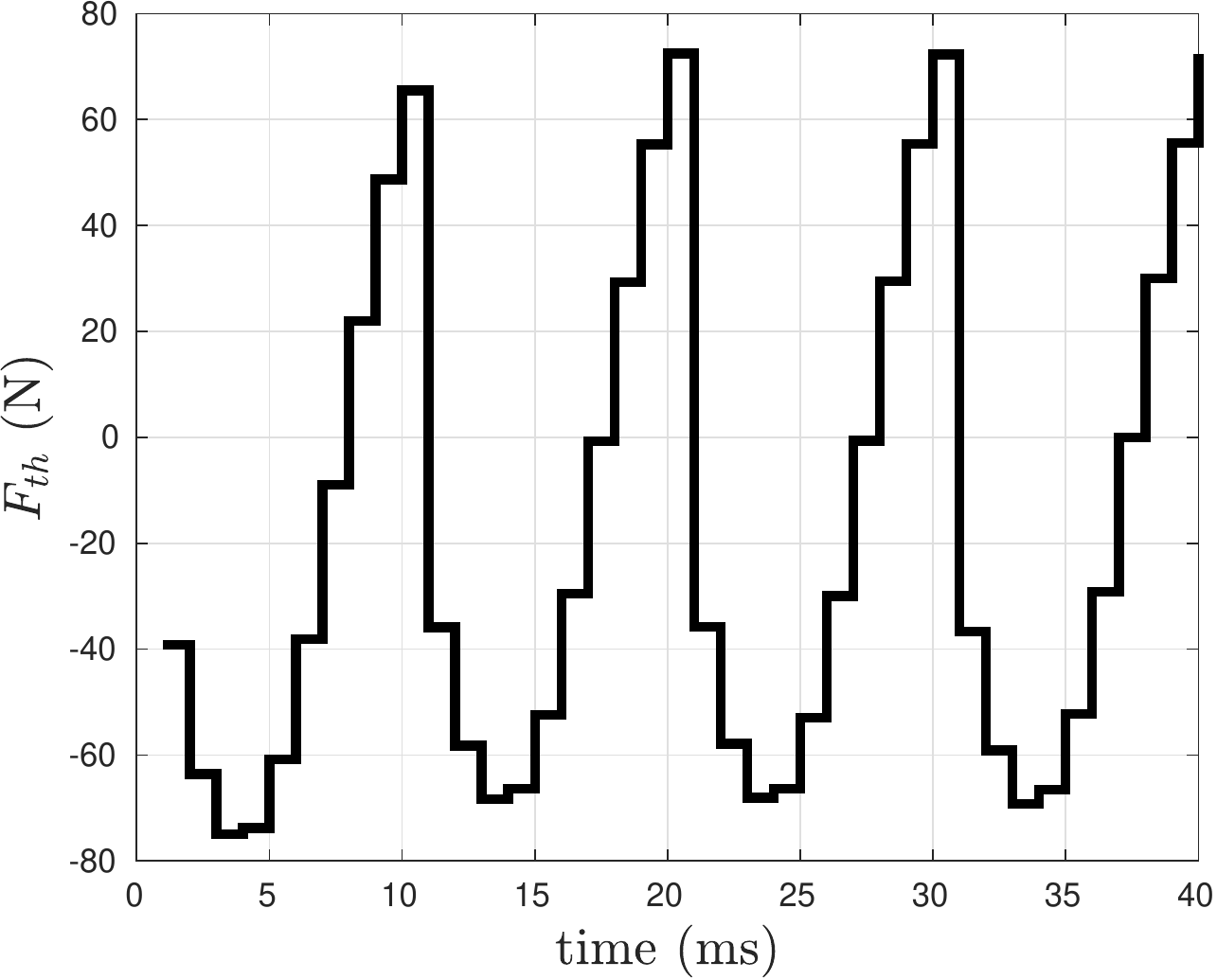}
    \caption{Thruster action during 10 ms DS phase. Intermediate SS phases were omitted as thrusters are inactive.}
    \label{fig:U_DS}
\end{figure}


\begin{figure}
\centering
    \includegraphics[width=0.9\linewidth]{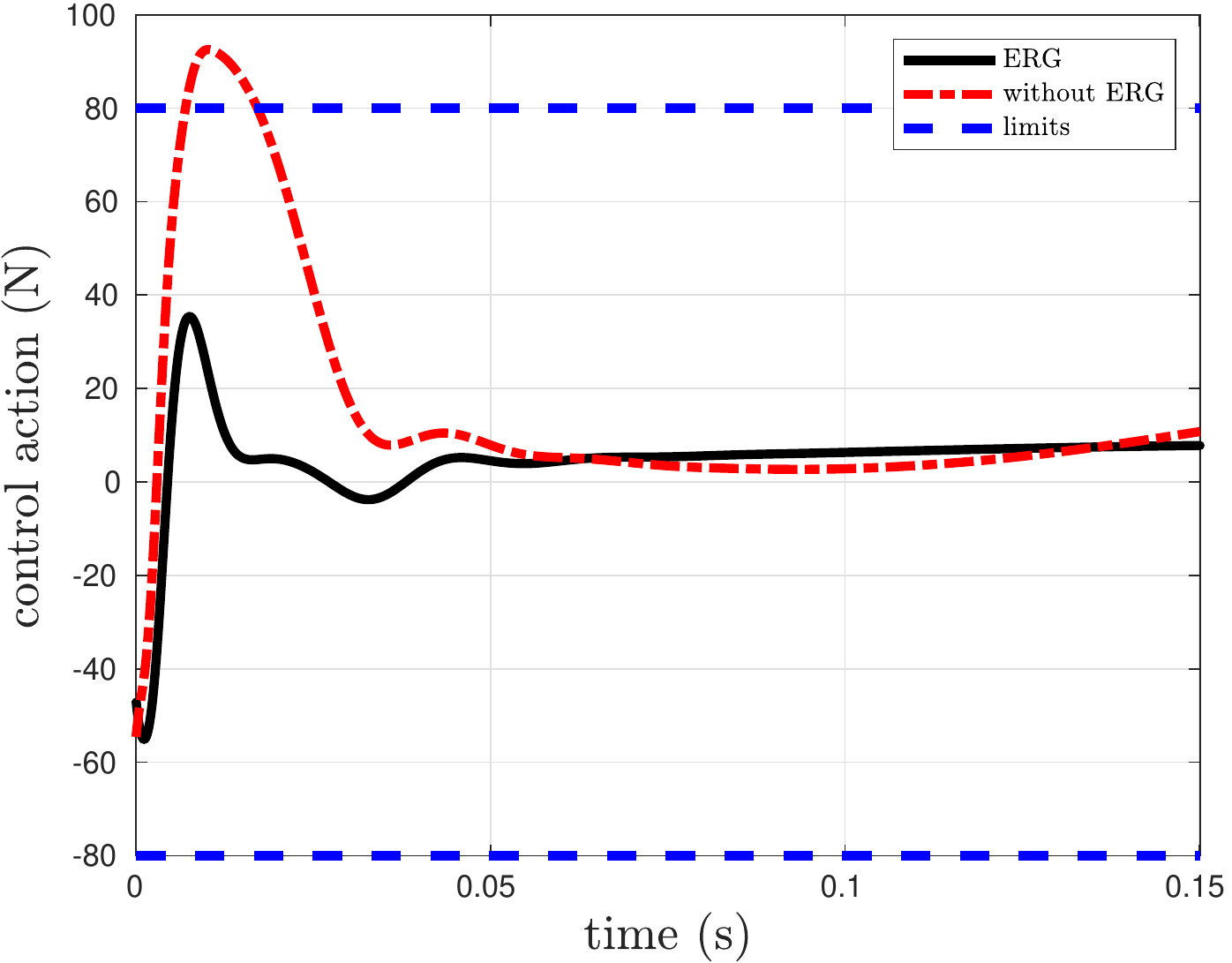}
\caption{Control input in the VLIP model when ERG applied versus standard control}
\label{fig:erg_u}
\end{figure}


\begin{figure}
\centering
\null\hfill
    \subfloat{
        \includegraphics[width=0.45\linewidth]{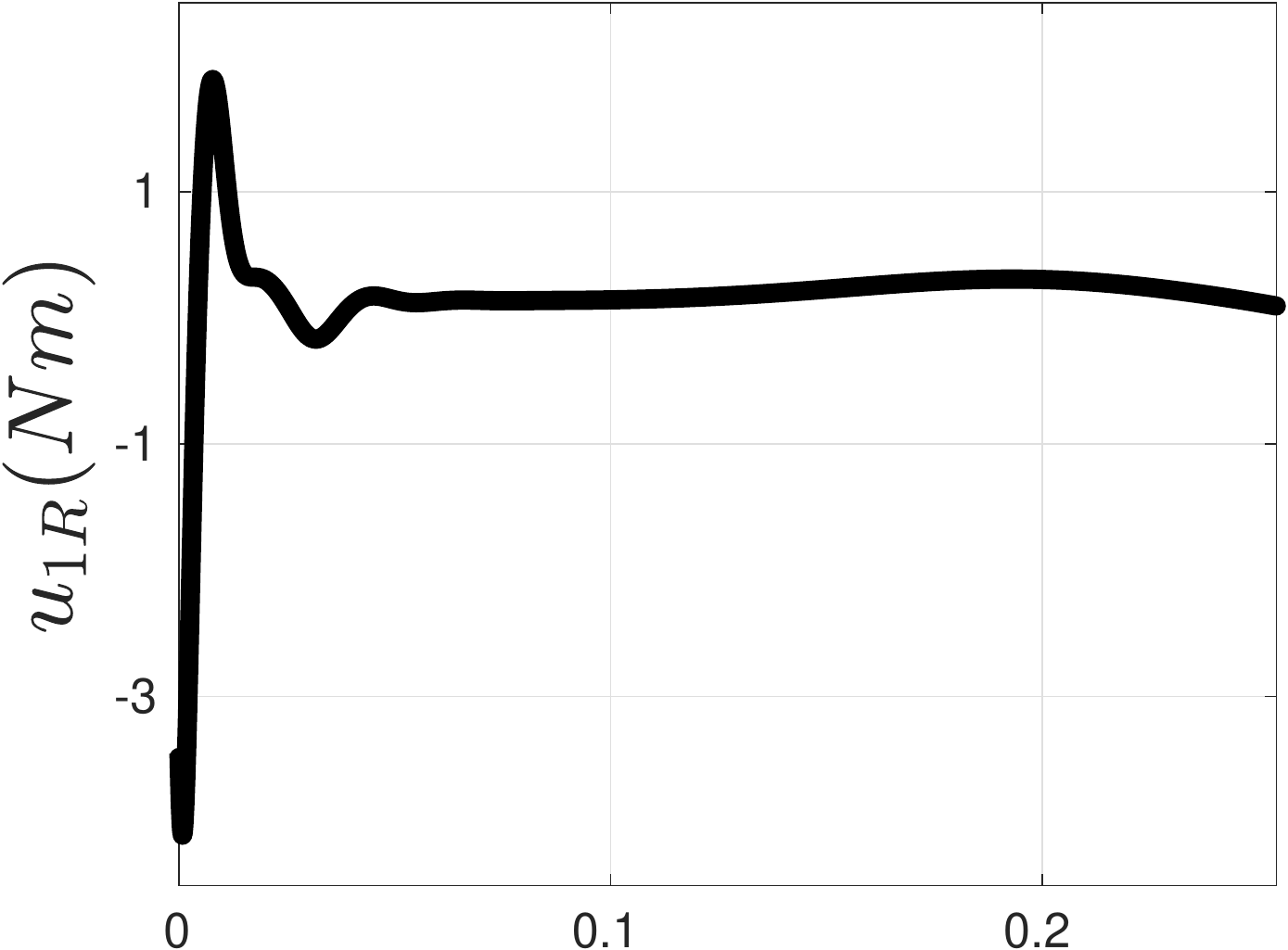} 
        }
    \hfill
    \subfloat[\label{fig:forces:u1L}]{
        \includegraphics[width=0.45\linewidth]{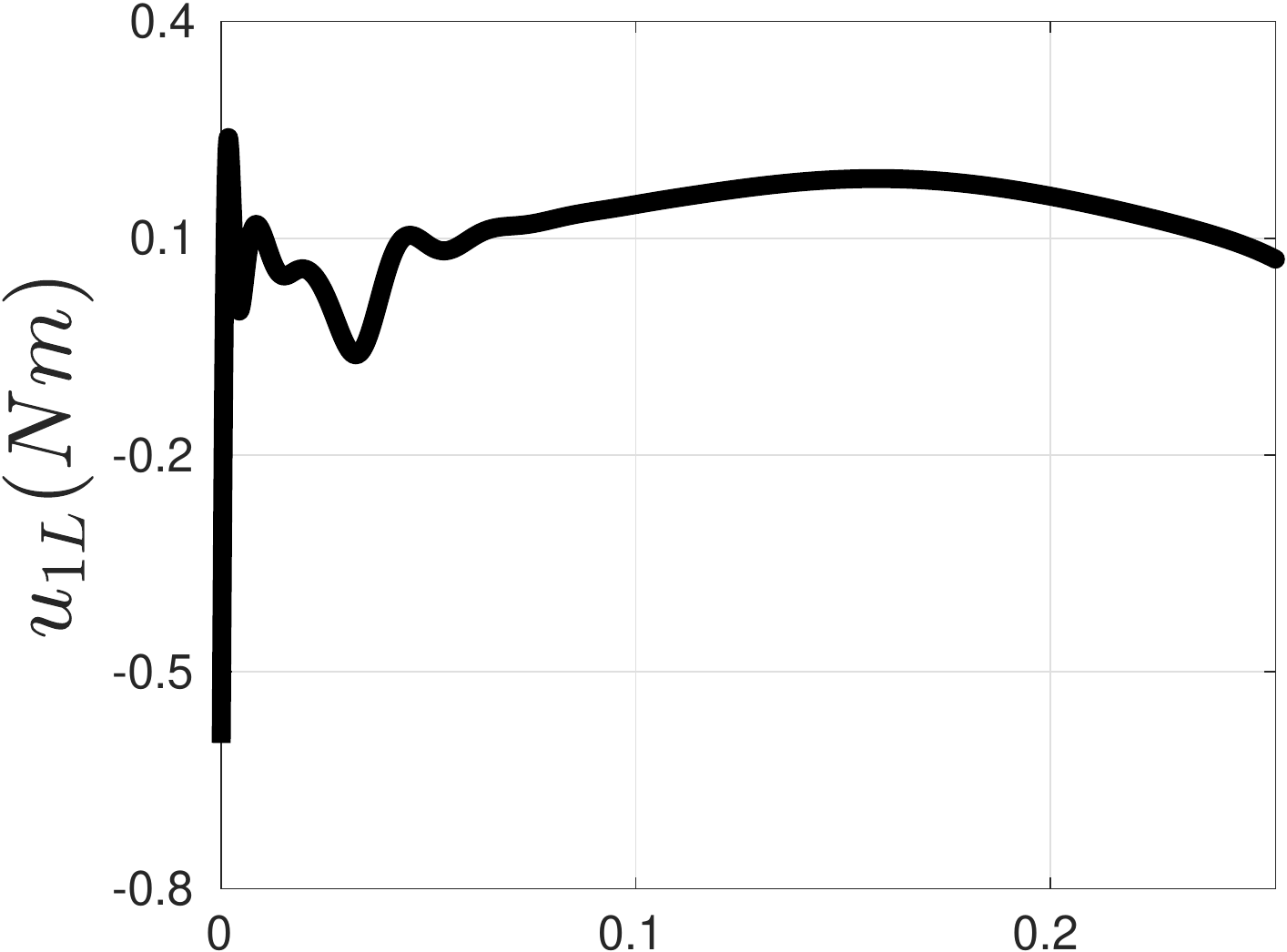}
        }
\hfill\null   
\hfill
\null\hfill
\subfloat{
        \includegraphics[width=0.45\linewidth]{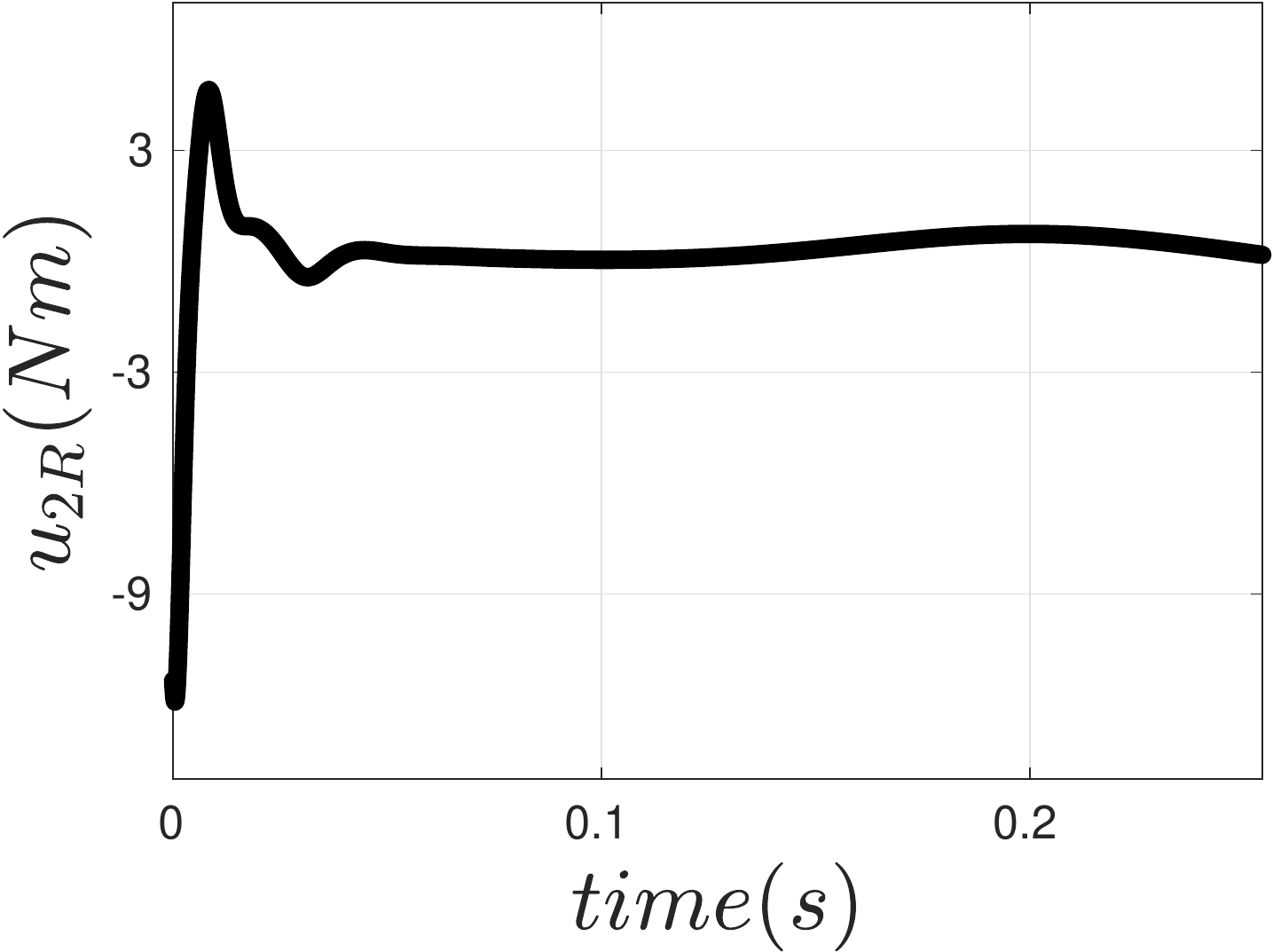}
        }
    \hfill
    \subfloat{
        \includegraphics[width=0.45\linewidth]{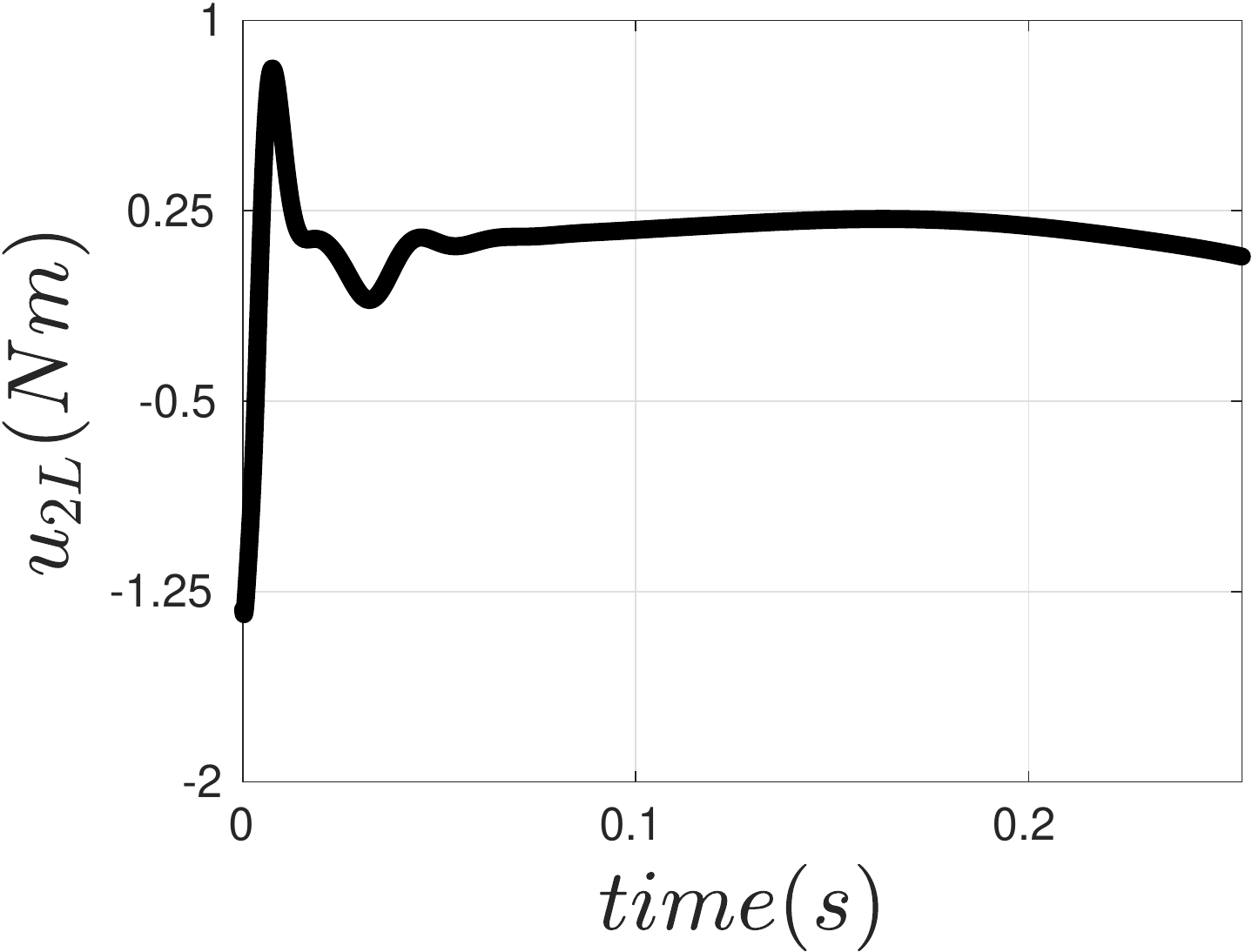}
        }
\hfill\null 
\caption{SS phase control actions mapped from VLIP model where ERG was applied}
\label{fig:u_mapped}
\end{figure}

A total of 5 stable steps were simulated to test the effectiveness of the proposed scheme on the planar hybrid model. The list of all model parameters are shown in table \ref{tab:modelparam}. For the SS phase the desired trajectory $h_d$ were parameterized as Bezier polynomials with its coefficients tuned offline such that states after two point impact are brought close as possible to initial SS phase states $x_{s,0}$. The resulting nominal COM position was then used as the reference trajectory for ERG. Each DS phase was simulated for a short fixed duration of 10 ms. 


Fig. \ref{fig:angles} and \ref{fig:velocity} show the periodic joint trajectories resulting from thruster assisted impact correction in DS phase. Fig. \ref{fig:PP_theta} shows a limit cycle on the phase portrait of horizontal hip position, here the SS phase is shown in black, straight green lines depicts impact and red lines indicate DS phase. It is seen that the effects of impact are corrected through correction made in DS phase.

The effects of thruster assisted locomotion is displayed in Fig. \ref{fig:forces}, which shows that the ground contact constraints are satisfied for each DS phase. In these figures intermediate SS phases are omitted and consecutive DS phases are stitched together. It should be noted that the normal forces exerted by leg 1 (Fig. \ref{fig:forces:normal1}) would not be achievable without the addition of thrusters (Fig. \ref{fig:U_DS}) as this would be limited by the total weight of the biped and inertial forces. The larger forces on leg 1, i.e. front leg, as opposed to leg 2 is due to it bearing the bulk of the load exerted by the weight and thrusters.

Fig. \ref{fig:erg_u} shows the consequence of ERG applied on the equivalent VLIP model. The bounded control action of the VLIP model are mapped on to SS phase actuators and it can be seen in Fig. \ref{fig:u_mapped} that the control actions are within achievable limits. 

Overall we can see from the results that the presence of the thruster, which provides an additional degree of actuation, allows for correction in DS phase necessary to achieve stable gaits. The combined control schemes utilized in SS and DS phase ensures the desired gaits are achieved without violating imposed actuator and ground contact constraints.


\begin{table} [ht!] \centering
    \begin{tabular}{l r l l} 
        \toprule
        Parameter &  Value & Description &  \\
        \midrule
        $m_T$ &  300 $g$ & Mass of torso &  \\
        $m_h$ &  200 $g$ & Mass of hip &  \\
        $m_k$ &  100 $g$ & Mass of each leg &  \\
        $l_T$ &  10 $cm$ & Length from hip to torso &  \\
        $l_1$ &  18 $cm$ & Length hip to knee &  \\
        $l_{2a}$ & 32 $cm$ & Length of tibia & \\ 
        $l_{2b}$ & 32 $cm$ & Length of metatarsus & \\ 
        \bottomrule
    \end{tabular}
    \caption{Model Parameters}
    \label{tab:modelparam}
\end{table}

 
\section {Conclusion} \label{section:conclusion}
This paper summarizes our recent efforts in designing feedback for the thruster-assisted walking of a bipedal robots. Firstly, gaits  were designed in SS phase following the well established HZD framework. To satisfy actuator constraint, an ERG method was used on an equivalent VLIP model to modify reference trajectories and the controller actions are mapped back to the full model. The modification in SS phase along with impact event were then mitigated by employing a predictive scheme which exploits the thrusters during DS phase leading to hybrid invariance. The combined efforts in SS and DS phase resulted in gaits that were stable and periodic.

\printbibliography

@book{westervelt2007feedback,
  title={Feedback Control of Dynamic Bipedal Robot Locomotion},
  author={Westervelt, E.R. and Grizzle, J.W.},
  isbn={9781420053722},
  lccn={2007007727},
  series={Control and Automation Series},
  url={https://books.google.com/books?id=xaMeAQAAIAAJ},
  year={2007},
  publisher={CRC PressINC}
}

@article{impact,
author = {Yildirim Hurmuzlu and Dan B. Marghitu},
title ={Rigid Body Collisions of Planar Kinematic Chains With Multiple Contact Points},
journal = {The International Journal of Robotics Research},
volume = {13},
number = {1},
pages = {82-92},
year = {1994},
doi = {10.1177/027836499401300106}
}

@book{khalil2002nonlinear,
  title={Nonlinear Systems},
  author={Khalil, H.K.},
  isbn={9780130673893},
  lccn={95045804},
  series={Pearson Education},
  url={https://books.google.com/books?id=t\_d1QgAACAAJ},
  year={2002},
  publisher={Prentice Hall}
}

@ARTICLE{bhat1998continuous,
author={S. P. {Bhat} and D. S. {Bernstein}},
journal={IEEE Transactions on Automatic Control},
title={Continuous finite-time stabilization of the translational and rotational double integrators},
year={1998},
volume={43},
number={5},
pages={678-682},
doi={10.1109/9.668834},
ISSN={0018-9286},
month=05,}

@ARTICLE{CLFQP,
author={K. {Galloway} and K. {Sreenath} and A. D. {Ames} and J. W. {Grizzle}},
journal={IEEE Access},
title={Torque Saturation in Bipedal Robotic Walking Through Control Lyapunov Function-Based Quadratic Programs},
year={2015},
volume={3},
number={},
pages={323-332},
doi={10.1109/ACCESS.2015.2419630},
ISSN={2169-3536},
month={03},}

@article{westervelt2003hybrid,
  title={Hybrid zero dynamics of planar biped walkers},
  author={Westervelt, Eric R and Grizzle, Jessy W and Koditschek, Daniel E},
  journal={IEEE transactions on automatic control},
  volume={48},
  number={1},
  pages={42--56},
  year={2003},
  publisher={IEEE}
}

@ARTICLE{898695,
author={J. W. {Grizzle} and G. {Abba} and F. {Plestan}},
journal={IEEE Transactions on Automatic Control},
title={Asymptotically stable walking for biped robots: analysis via systems with impulse effects},
year={2001},
volume={46},
number={1},
pages={51-64},
doi={10.1109/9.898695},
ISSN={0018-9286},
month={1},}

@article{choi_grizzle_2005, title={Feedback control of an underactuated planar bipedal robot with impulsive foot action}, volume={23}, DOI={10.1017/S0263574704001250}, number={5}, journal={Robotica}, publisher={Cambridge University Press}, author={Choi, Jun Ho and Grizzle, J. W.}, year={2005}, pages={567–580}}

@article{garone2015explicit,
  title={Explicit reference governor for constrained nonlinear systems},
  author={Garone, Emanuele and Nicotra, Marco M},
  journal={IEEE Transactions on Automatic Control},
  volume={61},
  number={5},
  pages={1379--1384},
  year={2015},
  publisher={IEEE}
}

@article{bemporad1998reference,
  title={Reference governor for constrained nonlinear systems},
  author={Bemporad, Alberto},
  journal={IEEE Transactions on Automatic Control},
  volume={43},
  number={3},
  pages={415--419},
  year={1998},
  publisher={IEEE}
}

@INPROCEEDINGS{411031,
author={E. G. {Gilbert} and I. {Kolmanovsky} and {Kok Tin Tan}},
booktitle={Proceedings of 1994 33rd IEEE Conference on Decision and Control},
title={Nonlinear control of discrete-time linear systems with state and control constraints: a reference governor with global convergence properties},
year={1994},
volume={1},
number={},
pages={144-149 vol.1},
doi={10.1109/CDC.1994.411031},
ISSN={},
month={12},}

@article{gilbert2002nonlinear,
  title={Nonlinear tracking control in the presence of state and control constraints: a generalized reference governor},
  author={Gilbert, Elmer and Kolmanovsky, Ilya},
  journal={Automatica},
  volume={38},
  number={12},
  pages={2063--2073},
  year={2002},
  publisher={Elsevier}
}

@inproceedings{saglam2015meshing,
  title={Meshing hybrid zero dynamics for rough terrain walking},
  author={Saglam, Cenk Oguz and Byl, Katie},
  booktitle={2015 IEEE International Conference on Robotics and Automation (ICRA)},
  pages={5718--5725},
  year={2015},
  organization={IEEE}
}

@article{spong2007passivity,
  title={Passivity-based control of bipedal locomotion},
  author={Spong, Mark W and Holm, Jonathan K and Lee, Dongjun},
  journal={IEEE Robotics \& Automation Magazine},
  volume={14},
  number={2},
  pages={30--40},
  year={2007},
  publisher={IEEE}
}

@INPROCEEDINGS{7803333,
author={H. {Dai} and R. {Tedrake}},
booktitle={2016 IEEE-RAS 16th International Conference on Humanoid Robots (Humanoids)},
title={Planning robust walking motion on uneven terrain via convex optimization},
year={2016},
volume={},
number={},
pages={579-586},
doi={10.1109/HUMANOIDS.2016.7803333},
ISSN={},
month={11},}

@INPROCEEDINGS{7041347,
author={S. {Feng} and E. {Whitman} and X. {Xinjilefu} and C. G. {Atkeson}},
booktitle={2014 IEEE-RAS International Conference on Humanoid Robots},
title={Optimization based full body control for the atlas robot},
year={2014},
volume={},
number={},
pages={120-127},
doi={10.1109/HUMANOIDS.2014.7041347},
ISSN={},
month={11},}

@article{sontag1983lyapunov,
  title={A Lyapunov-like characterization of asymptotic controllability},
  author={Sontag, Eduardo D},
  journal={SIAM journal on control and optimization},
  volume={21},
  number={3},
  pages={462--471},
  year={1983},
  publisher={SIAM}
}

@INPROCEEDINGS{371031,
author={P. V. {Kokotovic} and M. {Krstic} and I. {Kanellakopoulos}},
booktitle={[1992] Proceedings of the 31st IEEE Conference on Decision and Control},
title={Backstepping to passivity: recursive design of adaptive systems},
year={1992},
volume={},
number={},
pages={3276-3280 vol.4},
doi={10.1109/CDC.1992.371031},
ISSN={},
month={12},}

@article{raibert1984experiments,
  title={Experiments in balance with a 3D one-legged hopping machine},
  author={Raibert, Marc H and Brown Jr, H Benjamin and Chepponis, Michael},
  journal={The International Journal of Robotics Research},
  volume={3},
  number={2},
  pages={75--92},
  year={1984},
  publisher={Sage Publications Sage CA: Thousand Oaks, CA}
}

@article{raibert2008bigdog,
  title={Bigdog, the rough-terrain quadruped robot},
  author={Raibert, Marc and Blankespoor, Kevin and Nelson, Gabriel and Playter, Rob},
  journal={IFAC Proceedings Volumes},
  volume={41},
  number={2},
  pages={10822--10825},
  year={2008},
  publisher={Elsevier}
}

@article{hirose2006honda,
  title={Honda humanoid robots development},
  author={Hirose, Masato and Ogawa, Kenichi},
  journal={Philosophical Transactions of the Royal Society A: Mathematical, Physical and Engineering Sciences},
  volume={365},
  number={1850},
  pages={11--19},
  year={2006},
  publisher={The Royal Society London}
}

@inproceedings{kwon2007biped,
  title={Biped humanoid robot Mahru III},
  author={Kwon, Woong and Kim, Hyun K and Park, Joong Kyung and Roh, Chang Hyun and Lee, Jawoo and Park, Jaeho and Kim, Won-Kuk and Roh, Kyungshik},
  booktitle={2007 7th IEEE-RAS International Conference on Humanoid Robots},
  pages={583--588},
  year={2007},
  organization={IEEE}
}

@INPROCEEDINGS{5354430,
author={J. E. {Pratt} and B. {Krupp} and V. {Ragusila} and J. {Rebula} and T. {Koolen} and N. {van Nieuwenhuizen} and C. {Shake} and T. {Craig} and J. {Taylor} and G. {Watkins} and P. {Neuhaus} and M. {Johnson} and S. {Shooter} and K. {Buffinton} and F. {Canas} and J. {Carff} and W. {Howell}},
booktitle={2009 IEEE/RSJ International Conference on Intelligent Robots and Systems},
title={The Yobotics-IHMC Lower Body Humanoid Robot},
year={2009},
volume={},
number={},
pages={410-411},
doi={10.1109/IROS.2009.5354430},
ISSN={},
month={10},}

@article{chevallereau2004, title={Tracking a joint path for the walk of an underactuated biped}, volume={22}, DOI={10.1017/S0263574703005460}, number={1}, journal={Robotica}, publisher={Cambridge University Press}, author={Chevallereau, Christine and Formal'sky, Alexander and Djoudi, Dalila}, year={2004}, pages={15–28}}

@INPROCEEDINGS{1641816,
author={ {Guobiao Song} and M. {Zefran}},
booktitle={Proceedings 2006 IEEE International Conference on Robotics and Automation, 2006. ICRA 2006.},
title={Underactuated dynamic three-dimensional bipedal walking},
year={2006},
volume={},
number={},
pages={854-859},
doi={10.1109/ROBOT.2006.1641816},
ISSN={},
month={05},}

@INPROCEEDINGS{6225272,
author={G. {Garofalo} and C. {Ott} and A. {Albu-Schäffer}},
booktitle={2012 IEEE International Conference on Robotics and Automation},
title={Walking control of fully actuated robots based on the Bipedal SLIP model},
year={2012},
volume={},
number={},
pages={1456-1463},
doi={10.1109/ICRA.2012.6225272},
ISSN={},
month={05},}

\end{document}